\documentclass[letterpaper, 10 pt, conference]{ieeeconf}  

\IEEEoverridecommandlockouts                              

\usepackage{graphics} 
\usepackage{graphicx}
\usepackage{algorithm}
\usepackage{pgfplots}
\usepackage{algorithmic}
\usepackage{amsmath} 
\usepackage{amssymb}  
\usepackage[font={small,it}]{caption}
\usepackage{subcaption}
\usepackage{tikz}
\usetikzlibrary{positioning, shapes.geometric, arrows}

\tikzstyle{state}=[thick,draw=black,circle,minimum size = 7mm,inner sep=0mm]
\tikzstyle{action}=[draw=black,fill=black,circle,inner sep=0mm, minimum size=1.5mm]
\tikzstyle{arrow} = [thick,->]
\tikzstyle{process} = [rectangle, minimum width=0.9cm, minimum height=0.9cm, text centered, draw=black]

\title{\LARGE \bf
Stochastic Grounded Action Transformation for Robot Learning in Simulation
}

\author{Siddharth Desai$^{1\S}$, Haresh Karnan$^{1\S}$, Josiah P. Hanna$^{2}$, Garrett Warnell$^{3}$ and Peter Stone$^{4}$
\thanks{$^\S$Equal contribution}%
\thanks{$^{1}$ The University of Texas at Austin, Department of Mechanical Engineering
        {\tt\footnotesize \{sidrdesai,haresh.miriyala\}@utexas.edu}}%
\thanks{$^{2}$School of Informatics, University of Edinburgh; To be joining the Computer Sciences Department, University of Wisconsin---Madison
        {\tt\footnotesize josiah.hanna@ed.ac.uk}}%
\thanks{$^{3}$Army Research Laboratory
        {\tt\footnotesize garrett.a.warnell.civ@mail.mil}}%
\thanks{$^{4}$The University of Texas at Austin, Department of Computer Science and Sony AI
        {\tt\footnotesize pstone@cs.utexas.edu}}%
}

\begin{document}

\maketitle
\thispagestyle{empty}
\pagestyle{empty}

\begin{abstract}


Robot control policies learned in simulation do not often transfer well to the real world.
Many existing solutions to this {\em sim-to-real} problem, such as the Grounded Action Transformation (\textsc{gat}) algorithm, seek to correct for---or {\em ground}---these differences by matching the simulator to the real world.
However, the efficacy of these approaches is limited if they do not explicitly account for stochasticity in the target environment.
In this work, we analyze the problems associated with grounding a deterministic simulator in a stochastic real world environment, and we present examples where \textsc{gat} fails to transfer a good policy due to stochastic transitions in the target domain.
In response, we introduce the {\em Stochastic Grounded Action Transformation} (\textsc{sgat}) algorithm, which models this stochasticity when grounding the simulator.
We find experimentally---for both simulated and physical target domains---that \textsc{sgat} can find policies that are robust to stochasticity in the target domain.
\end{abstract}

\section{INTRODUCTION}

Learning robot control policies in simulation \cite{Cutler2015EfficientRL} is typically safer, cheaper, and faster than learning in the real world, but it also introduces a \textit{reality gap} \cite{Kober2012ReinforcementLI} between the training environment (the simulator) and the deployment environment (the real world). {\em Sim-to-real} algorithms,  which focus on overcoming this gap,  have recently received a great deal of attention.

This paper focuses on a class of solutions referred to as grounding algorithms \cite{AAMAS13-Farchy} which use a small amount of real world data to improve (or \textit{ground}) the simulator.
We can assume that the ungrounded simulator at least approximates the correct real world dynamics, so many grounding approaches \cite{AAMAS13-Farchy, Tan_2018, nas_golemo, AAAI17-Hanna} learn parameters to correct for these differences.
Since it may not always be feasible to modify the internal parameters of a simulator, we focus our attention on grounding algorithms that treat the simulator as a black-box, such as the Grounded Action Transformation (\textsc{gat}) algorithm \cite{AAAI17-Hanna}.

The dynamics of robots are often best modeled as stochastic processes due to effects like friction, gear backlash, uneven terrain, and other sources of noise in the environment; however many earlier sim-to-real approaches do not explicitly account for stochasticity in the real world. These approaches learn a single value for the correction terms whereas learning a distribution would more accurately reflect the real world.

We hypothesize that accounting for the stochasticity in the real world improves sim-to-real transfer of policies learned in simulation. In this work, we analyze the effects of target environment stochasticity in the {\em sim-to-real} problem. We show several domains where \textsc{gat} fails to adequately ground the simulator, and we propose a new algorithm, {\em Stochastic Grounded Action Transformation} (\textsc{sgat}), that handles this issue gracefully by learning the stochasticity in the environment.
We first show on the \textit{Cliff Walking} domain that using \textsc{sgat} achieves better mean returns than \textsc{gat} on the target environment. We further conduct sim-to-``real" transfer experiments on OpenAI gym MuJoCo environments \textit{InvertedPendulum} and \textit{HalfCheetah}, using a modified version of the environment as a surrogate for the real world.

\begin{figure*}[!tb]
    \centering
    \includegraphics[scale=1.1, trim={0 0.3cm 0 0.3cm} ,clip]{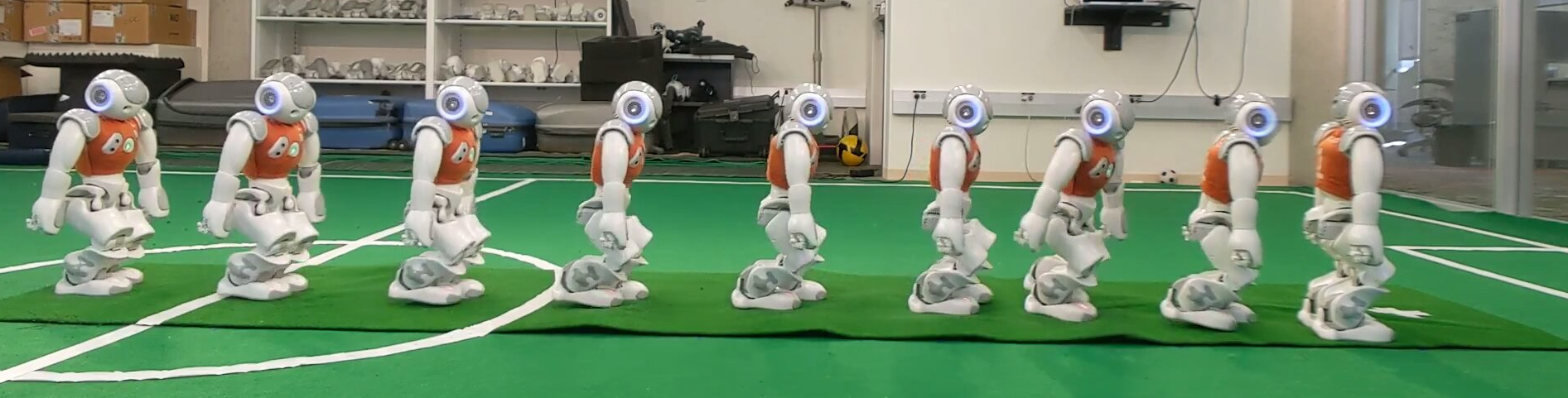}
    \caption{Experiment setup showing a  robot walking on the uneven ground. The \textsc{nao} begins walking 40 cms behind the center of the circle and walks 300 cms towards the white penalty cross. This image shows a successful walk executed by the robot at 2 sec intervals, learned using the proposed \textsc{sgat} algorithm.}
    \label{fig:lumpy_field_setup}
\end{figure*}

To test the efficacy of \textsc{sgat} in handling real world stochasticity, we set up a sim-to-real experiment where a humanoid \textsc{nao} robot learns to walk on uneven ground. Confirming our hypothesis, we indeed found that policies trained with \textsc{sgat}, unlike those learned by \textsc{gat}, learned to cope with environment stochasticity, leading to better real world performance. Using \textsc{sgat}, the \textsc{nao} completed the course 9 out of 10 times (Fig. \ref{fig:lumpy_field_setup}), whereas with \textsc{gat}, it fell down every time.

\section{BACKGROUND}

In this section, we review existing literature on state-of-the-art sim-to-real algorithms that are relevant to our approach, and we formalize the problem addressed in this paper.

\subsection{Related Work}

In the past, many sim-to-real techniques \cite{Tan_2018, DBLP:journals/corr/abs-1710-06537, Jakobi1995, miglino1995, rajeswaran2016epopt} have achieved policy learning for the real world using a simulator.
We can roughly divide these techniques into black-box techniques (which do not modify the simulator internally) and white-box techniques (which do). Another useful distinction is robustness methods vs. grounding methods.

Robustness methods learn a policy in simulation that is robust to variations in the environment. Dynamics Randomization (\textsc{DR}) is a relatively simple approach that has been shown to be successful at robot control tasks.
Peng et al. \cite{DBLP:journals/corr/abs-1710-06537} introduce a white-box \textsc{DR} algorithm that can learn robust RL control policies for robotic manipulation tasks.
DR approaches where the effect of actuators (actions) or sensor information (observations) are perturbed with an ``envelope" of noise can be considered black-box approaches \cite{Jakobi1995, miglino1995}. Other DR methods where the internal simulation parameters are perturbed fall under white-box approaches \cite{Tan_2018, DBLP:journals/corr/abs-1710-06537, Molchanov_2019, Yu_2017}. Robust Adversarial Reinforcement Learning (\textsc{rarl}) \cite{pinto2017robust} randomizes the training environment using an adversarial agent. While DR methods randomly sample simulation parameters before a trajectory is generated, the simulator dynamics may still be deterministic. Our approach models real world stochasticity during individual state transitions.

Orthogonal to robustness methods are approaches that focus on grounding a simulator to behave like a target real world environment.
Unlike DR methods that learn a more general robust policy to perform well in any environment, grounding methods optimize performance in that specific target environment. Grounded Simulation Learning (\textsc{gsl}) \cite{AAMAS13-Farchy} is a learning framework in which data from the real world is used to modify (ground) the simulator. After this grounding step, transitions in the grounded simulator look similar to transitions in the real world environment, hence minimizing the ``reality gap". In \textsc{gsl}, all of the policy learning happens in the simulator; the robot is used only to evaluate policies and to collect transition data. Farchy et al. demonstrate its success at transferring a walk policy from a simulated \textsc{nao} in the SimSpark simulator to a real SoftBank \textsc{nao} and achieved 26.7\% faster walk than baseline methods  \cite{AAMAS13-Farchy}.

Grounded Action Transformation (\textsc{gat}) \cite{AAAI17-Hanna} is a recent black-box algorithm based on the \textsc{gsl} framework.
\textsc{gat} was experimentally shown to be effective at learning a fast walk policy by grounding a high fidelity simulator, which when deployed in the real world achieved the fastest known walk on the \textsc{nao} robot: over 43\% faster than the state-of-the-art hand-coded gait upon which it was based.

\subsection{Preliminaries}

We consider the real world (real) and simulation (sim) domains to be two different Markov Decision Processes (\textsc{mdp}s) \cite{Suttonbook}, $\mathcal{M}_{real}$ and $\mathcal{M}_{sim}$ respectively. An \textsc{mdp} comprises a set of states, $\mathcal{S}$, a set of actions, $\mathcal{A}$, the transition dynamics associated with those actions, $\mathcal{T}$, and a reward function, $R$. At each time step, $t$, an agent observes the current state, $s_t \in \mathcal{S}$, and chooses an action, $a_t \in \mathcal{A}$, sampled from its policy, $a_t \sim \pi(\cdot|s_t)$. 
For a given state and action, there is a distribution of possible next states, from which the environment samples a state---$s_{t+1} \sim \mathcal{T}(\cdot|s_t,a_t)$. 
The reward, in this work, is a function of the action and the next state, $r_{t+1} = R(a_t, s_{t+1})$.

In our formulation of the sim-to-real problem, only the transition dynamics $\mathcal{T}_{sim}$ and $\mathcal{T}_{real}$ differ between the environments. Furthermore, we only consider deterministic simulators in this work, but the real environment may have stochastic transitions. The reinforcement learning (RL) objective is to find a policy that maximizes the expected sum of rewards on the real domain, $\mathbb{E}[\sum_{t=0}^T R(a_t, s_{t+1})]$.

\subsection{Grounded Action Transformation (\textsc{gat})}

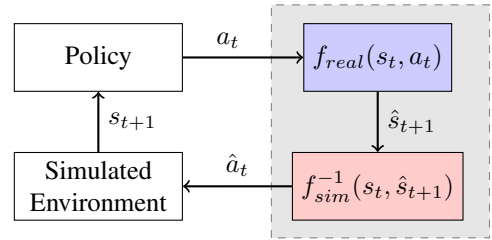
\begin{figure}[!b]
    \centering
    \begin{tikzpicture}
    \node[process, text width=20mm] (pol) {Policy};
    \filldraw[fill=black!10, draw=black!50, dashed] (2.3,-2.4) rectangle (5.2,.7);
    \node[process, right=16mm of pol, fill=blue!20] (fwd) {$f_{real}(s_t,a_t)$};
    \node[process, below=8mm of fwd, fill=red!20] (inv) {$f^{-1}_{sim}(s_t,\hat{s}_{t+1})$};
    \node[process, below=8mm of pol, text width=20mm] (env) {Simulated Environment};
    \draw[arrow] (pol) -- node [above, xshift=-2mm] {$a_t$} (fwd);
    \draw[arrow] (fwd) -- node [right] {$\hat{s}_{t+1}$} (inv);
    \draw[arrow] (inv) -- node [above] {$\hat{a}_t$} (env);
    \draw[arrow] (env) -- node[right] {$s_{t+1}$} (pol);
    \end{tikzpicture}
    \caption{\textsc{gat} Diagram\cite{AAAI17-Hanna}}
    \label{fig:gat_diagram}
\end{figure}

\textsc{gat} follows the \textsc{gsl} framework, alternating between a \textit{grounding step} and a \textit{policy improvement step}. During the grounding step, the policy remains unchanged and is deployed on both sim and real environments to collect state transition data. The \textsc{gat} algorithm trains two neural networks---a deterministic forward model $\hat{s}_{t+1}=f_{real}(s_t,a_t)$ and an inverse dynamics model $\hat{a}_t=f^{-1}_{sim}(s_t,{s}_{t+1})$  of the real and sim environments respectively, using supervised learning. The learned forward and inverse models are composed to form the action transformer function $g(s_t,a_t)=f^{-1}_{sim}(s_t,f_{real}(s_t,a_t))$ that grounds the simulator. The inputs to the action transformer are the current state, $s_t$, and the action chosen by the agent, $a_t \sim \pi(\cdot|s_t)$. The output is a transformed action, $\hat{a}_t$, that when taken in the simulator, produces a transition similar to the expected transition when executing $a_t$ in the real world. The combination of the action transformer and the original simulator is known as the grounded simulator. The grounding step is followed by a policy improvement step, in which the parameters of the action transformer are fixed. Thus, the policy improvement step is a standard reinforcement learning problem in the grounded simulator. These two steps are repeated until the optimized policy performs well on the real environment. The block diagram depicting the \textsc{gat} framework is shown in Fig. \ref{fig:gat_diagram}. While \textsc{gat} works well on fairly deterministic environments, as was shown by Hanna and Stone \cite{AAAI17-Hanna}, in our experimentation, we find that policies learned using \textsc{gat} perform poorly when transferring to highly stochastic environments.

\subsection{Transferring to a Stochastic Real World}

\begin{figure}[!tb]
\centering
\begin{subfigure}{.2\textwidth}
    \centering
    \begin{tikzpicture}[auto,>=latex]
        \node[state] (s0) {$s_0$};
        \node[action,above right=2mm and 4mm of s0,label={$a_1$}] (a1) {};
        \node[action,below right=2mm and 4mm of s0,label={$a_2$}] (a2) {};
        \draw[arrow] (s0) -- (a1);
        \draw[arrow] (s0) -- (a2);
        \node[state, right=8mm of a1, label=right:{+1}] (s1) {$s_1$};
        \node[state, right=8mm of a2, label=right:{-1}] (s2) {$s_2$};
        \draw[arrow] (a1) -- (s1);
        \draw[arrow] (a2) -- (s2);
    \end{tikzpicture}
    \caption{Simulator}
    \label{fig:toy1a}
\end{subfigure}
\begin{subfigure}{.2\textwidth}
    \centering
    \begin{tikzpicture}[auto,>=latex]
        \node[state] (s0) {$s_0$};
        \node[action,above right=2mm and 4mm of s0,label={$a_1$}] (a1) {};
        \node[action,below right=2mm and 4mm of s0,label={$a_2$}] (a2) {};
        \draw[arrow] (s0) -- (a1);
        \draw[arrow] (s0) -- (a2);
        \node[state, right=8mm of a1, label=right:{+1}] (s1) {$s_1$};
        \node[state, right=8mm of a2, label=right:{-1}] (s2) {$s_2$};
        \draw[arrow] (a1) -- (s2);
        \draw[arrow] (a2) -- (s1);
    \end{tikzpicture}
    \caption{Real World}
    \label{fig:toy1b}
\end{subfigure}
\begin{subfigure}{.4\textwidth}
    \centering
    \begin{tikzpicture}[auto,>=latex]
        \node[state] (s0) {$s_0$};
        \filldraw[fill=black!10, draw=black!50, dashed] (0.5,-1) rectangle (3,1);
        \node[action,above right=2mm and 4mm of s0,label={$a_1$}] (a1) {};
        \node[action,below right=2mm and 4mm of s0,label={$a_2$}] (a2) {};
        \draw[arrow] (s0) -- (a1);
        \draw[arrow] (s0) -- (a2);
        \node[state, dashed, right=8mm of a1] (s1i) {$\hat{s}_1$};
        \node[state, dashed, right=8mm of a2] (s2i) {$\hat{s}_2$};
        \draw[arrow, color=blue] (a1) -- (s2i);
        \draw[arrow, color=blue] (a2) -- (s1i);
        \node[action, right=8mm of s1i,label={$\hat{a}_1$}] (a1i) {};
        \node[action, right=8mm of s2i,label={$\hat{a}_2$}] (a2i) {};
        \draw[arrow, color={red}] (s1i) -- (a1i);
        \draw[arrow, color={red}] (s2i) -- (a2i);
        \node[state, right=8mm of a1i, label=right:{+1}] (s1) {$s_1$};
        \node[state, right=8mm of a2i, label=right:{-1}] (s2) {$s_2$};
        \draw[arrow] (a1i) -- (s1);
        \draw[arrow] (a2i) -- (s2);
    \end{tikzpicture}
    \caption{\textsc{gat} Grounded Simulator}
    \label{fig:toy1c}
\end{subfigure}
\caption{For a deterministic domain, the forward model (blue) and the inverse model (red) can make the grounded simulator behave like the real world. \textsc{gat} works well here.}
\label{fig:toy1}
\end{figure}
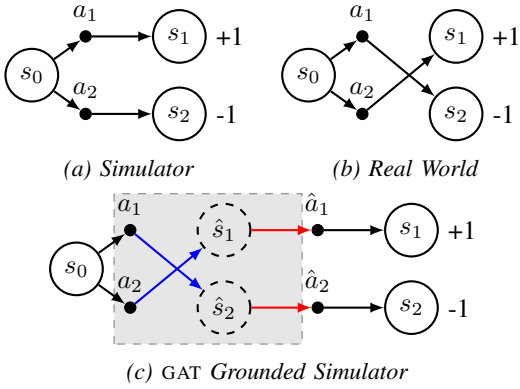

When the real world domain is deterministic, learning a deterministic forward model, as \textsc{gat} does, works well.
We can demonstrate using a toy example. Consider the environments shown in Figs. \ref{fig:toy1a} and \ref{fig:toy1b}. The agent starts in the initial state, $s_0$, and chooses between $a_1$ and $a_2$. In the simulator, $a_1$ leads to the higher reward of +1, but the transitions are flipped in reality. Hence, the action $a_2$ is the optimal action. The \textsc{gat} action transformer learns to transform $a_1$ into $a_2$ and $a_2$ into $a_1$. Thus from the agent's perspective the grounded simulator behaves like the real world. 

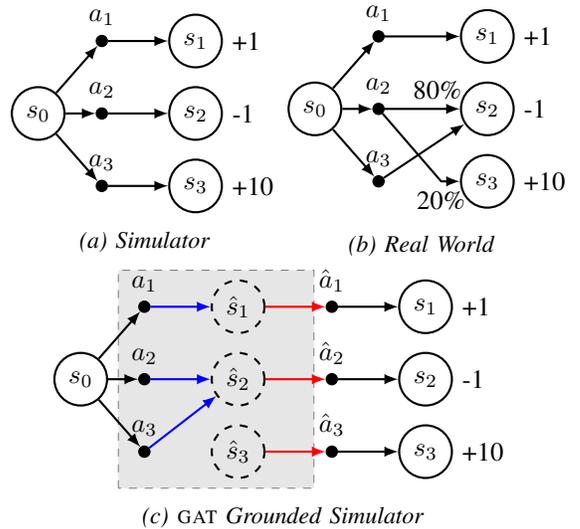
\begin{figure}[!tb]
\centering
\begin{subfigure}{.2\textwidth}
    \centering
    \begin{tikzpicture}[auto,>=latex]
        \node[state] (s0) {$s_0$};
        \node[action, right=4mm of s0,label={$a_2$}] (a2) {};
        \node[action, above=8mm of a2,label={$a_1$}] (a1) {};
        \node[action, below=8mm of a2,label={$a_3$}] (a3) {};
        \draw[arrow] (s0) -- (a1);
        \draw[arrow] (s0) -- (a2);
        \draw[arrow] (s0) -- (a3);
        \node[state, right=8mm of a1, label=right:{+1}] (s1) {$s_1$};
        \node[state, right=8mm of a2, label=right:{-1}] (s2) {$s_2$};
        \node[state, right=8mm of a3, label=right:{+10}] (s3) {$s_3$};
        \draw[arrow] (a1) -- (s1);
        \draw[arrow] (a2) -- (s2);
        \draw[arrow] (a3) -- (s3);
    \end{tikzpicture}
    \caption{Simulator}
    \label{fig:toy2a}
\end{subfigure}
\begin{subfigure}{.2\textwidth}
    \centering
    \begin{tikzpicture}[auto,>=latex]
        \node[state] (s0) {$s_0$};
        \node[action, right=4mm of s0,label={$a_2$}] (a2) {};
        \node[action, above=8mm of a2,label={$a_1$}] (a1) {};
        \node[action, below=8mm of a2,label={$a_3$}] (a3) {};
        \draw[arrow] (s0) -- (a1);
        \draw[arrow] (s0) -- (a2);
        \draw[arrow] (s0) -- (a3);
        \node[state, right=10mm of a1, label=right:{+1}] (s1) {$s_1$};
        \node[state, right=10mm of a2, label=right:{-1}] (s2) {$s_2$};
        \node[state, right=10mm of a3, label=right:{+10}] (s3) {$s_3$};
        \coordinate [left=2.5mm of s3, label=below:{20\%}] (c3);
        \draw[arrow] (a1) -- (s1);
        \draw[arrow] (a2) -- node [above, xshift=2mm] {80\%} (s2);
        \draw[arrow] (a2) -- (c3) -- (s3);
        \draw[arrow] (a3) -- (s2);
    \end{tikzpicture}
    \caption{Real World}
    \label{fig:toy2b}
\end{subfigure}
\begin{subfigure}{.4\textwidth}
    \centering
    \begin{tikzpicture}[auto,>=latex]
        \node[state] (s0) {$s_0$};
        \filldraw[fill=black!10, draw=black!50, dashed] (0.5,-1.45) rectangle (3.1,1.45);
        \node[action, right=4mm of s0,label={$a_2$}] (a2) {};
        \node[action, above=8mm of a2,label={$a_1$}] (a1) {};
        \node[action, below=8mm of a2,label={$a_3$}] (a3) {};
        \draw[arrow] (s0) -- (a1);
        \draw[arrow] (s0) -- (a2);
        \draw[arrow] (s0) -- (a3);
        \node[state, dashed, right=8mm of a1] (s1i) {$\hat{s}_1$};
        \node[state, dashed, right=8mm of a2] (s2i) {$\hat{s}_2$};
        \node[state, dashed, right=8mm of a3] (s3i) {$\hat{s}_3$};
        \draw[arrow, color=blue] (a1) -- (s1i);
        \draw[arrow, color=blue] (a2) -- (s2i);
        \draw[arrow, color=blue] (a3) -- (s2i);
        \node[action, right=8mm of s1i,label={$\hat{a}_1$}] (a1i) {};
        \node[action, right=8mm of s2i,label={$\hat{a}_2$}] (a2i) {};
        \node[action, right=8mm of s3i,label={$\hat{a}_3$}] (a3i) {};
        \draw[arrow, color=red] (s1i) -- (a1i);
        \draw[arrow, color=red] (s2i) -- (a2i);
        \draw[arrow, color=red] (s3i) -- (a3i);
        \node[state, right=8mm of a1i, label=right:{+1}] (s1) {$s_1$};
        \node[state, right=8mm of a2i, label=right:{-1}] (s2) {$s_2$};
        \node[state, right=8mm of a3i, label=right:{+10}] (s3) {$s_3$};
        \draw[arrow] (a1i) -- (s1);
        \draw[arrow] (a2i) -- (s2);
        \draw[arrow] (a3i) -- (s3);
    \end{tikzpicture}
    \caption{\textsc{gat} Grounded Simulator}
    \label{fig:toy2c}
\end{subfigure}
\caption{When the real world has stochastic transitions, the \textsc{gat} forward model (blue) only captures the most likely next state. \textsc{gat} may fail here, since the grounded simulator behaves very differently from the real environment.}
\label{fig:toy2}
\end{figure}

However, when we add stochastic transitions, the two diagrams do not match. In Fig. \ref{fig:toy2}, the optimal action in the simulator is $a_3$, and in the real world, it is $a_2$; however, in the grounded simulator, it is $a_1$.  Since \textsc{gat}'s forward model is deterministic, it predicts only the most likely next state, but other, less likely transitions are also important when computing an action's value.

\section{STOCHASTIC GROUNDED ACTION TRANSFORMATION (\textsc{SGAT})}

\begin{figure}[!tb]
    \centering
    \begin{tikzpicture}[auto,>=latex]
        \node[state] (s0) {$s_0$};
        \filldraw[fill=black!10, draw=black!50, dashed] (0.5,-1.45) rectangle (3.1,1.45);
        \node[action, right=4mm of s0,label={$a_2$}] (a2) {};
        \node[action, above=8mm of a2,label={$a_1$}] (a1) {};
        \node[action, below=8mm of a2,label={$a_3$}] (a3) {};
        \draw[arrow] (s0) -- (a1);
        \draw[arrow] (s0) -- (a2);
        \draw[arrow] (s0) -- (a3);
        \node[state, dashed, right=8mm of a1] (s1i) {$\hat{s}_1$};
        \node[state, dashed, right=8mm of a2] (s2i) {$\hat{s}_2$};
        \node[state, dashed, right=8mm of a3] (s3i) {$\hat{s}_3$};
        \coordinate [left=2.5mm of s3i, label=below:{20\%}] (c3);
        \draw[arrow, color=blue] (a1) -- (s1i);
        \draw[arrow, color=blue] (a2) -- node [above, xshift=2mm, color=black] {80\%} (s2i);
        \draw[arrow, color=blue] (a2) -- (c3) -- (s3i);
        \draw[arrow, color=blue] (a3) -- (s2i);
        \node[action, right=8mm of s1i,label={$\hat{a}_1$}] (a1i) {};
        \node[action, right=8mm of s2i,label={$\hat{a}_2$}] (a2i) {};
        \node[action, right=8mm of s3i,label={$\hat{a}_3$}] (a3i) {};
        \draw[arrow, color=red] (s1i) -- (a1i);
        \draw[arrow, color=red] (s2i) -- (a2i);
        \draw[arrow, color=red] (s3i) -- (a3i);
        \node[state, right=8mm of a1i, label=right:{+1}] (s1) {$s_1$};
        \node[state, right=8mm of a2i, label=right:{-1}] (s2) {$s_2$};
        \node[state, right=8mm of a3i, label=right:{+10}] (s3) {$s_3$};
        \draw[arrow] (a1i) -- (s1);
        \draw[arrow] (a2i) -- (s2);
        \draw[arrow] (a3i) -- (s3);
    \end{tikzpicture}
    \caption{In the \textsc{sgat} Grounded Simulator, the transitions match the real environment (Fig. \ref{fig:toy2b}).}
    \label{fig:toy3}
\end{figure}
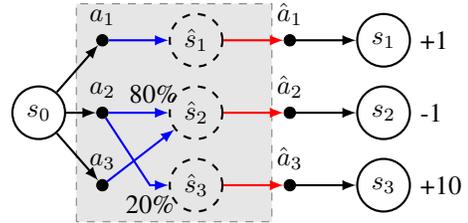

To address real world stochasticity, we introduce Stochastic Grounded Action Transformation (\textsc{sgat}), which learns a stochastic model of the forward dynamics. In other words, the learned forward model, $f_{real}$, predicts a distribution over next states (from which we sample) rather than the most likely next state. The sampling operation within the action transformer makes the overall process stochastic.
Fig. \ref{fig:toy3} illustrates the simulator from Fig. \ref{fig:toy2} grounded using \textsc{sgat}. Since the forward model accounts for stochasticity in the real world, the actions in the grounded simulator have the same effect as in the real world.

In continuous state and action domains, we model the next state as a multivariate Gaussian distribution and train the forward model using negative log likelihood (\textsc{nll}) loss $\mathcal{L} = -\log p(s_{t+1}|s_t,a_t)$. 
Similar to \textsc{gat}, we use a neural network function approximator with two fully connected hidden layers of 64 neurons to represent the forward and inverse models, but unlike \textsc{gat}, the forward model in \textsc{sgat} outputs the parameters of a Gaussian distribution from which we sample the predicted next state.\footnote{A Mixture Density Network might be more suitable when the environment's transition dynamics exhibit multimodal behavior.}
In our implementation, the final dense layer outputs the mean, $\mu$, and the log standard deviation, $log(\sigma)$, for each element of the state vector.
The complete algorithm of \textsc{sgat} is shown in Algorithm \ref{alg:SGAT}.

\begin{algorithm}[!tb]
\caption{Stochastic Grounded Action Transformation}
\label{alg:SGAT}
\textbf{Input}: initial parameters $\theta_0$, $\phi$, and $\psi$ for the policy $\pi_\theta$, forward dynamics model $f_{{\phi}_{real}}$ and inverse dynamics model $f^{-1}_{{\psi}_{sim}}$; policy improvement method, \texttt{optimize}
\begin{algorithmic}[1]
\WHILE{policy $\pi_\theta$ improves on real}
 \STATE Collect real world trajectories \\$\tau_{real} \leftarrow \{((s_0,a_0),s_1), ((s_1, a_1), s_2),...\}_{real}$
 \STATE Train forward dynamics function $f_\phi$ with $\tau_{real}$, using the \textsc{nll} loss $\mathcal{L} = -log$ $p(s_{t+1}|s_t,a_t)$.
 \STATE Collect simulated trajectories \\$\tau_{sim} \leftarrow \{((s_0,a_0),s_1), ((s_1, a_1), s_2),...\}_{sim}$
 \STATE Train inverse dynamics function $f^{-1}_\phi$ with $\tau_{sim}$, using mean squared error loss.
 \STATE Update $\pi_\theta$ on the grounded simulator using \texttt{optimize} and the reward from the grounded simulator
\ENDWHILE
\end{algorithmic}
\end{algorithm}

\section{EXPERIMENTS}

This section reports on an empirical study of transfer from simulation with \textsc{sgat} compared to \textsc{gat}.
We begin with a toy RL domain and progress to sim-to-real transfer of a bipedal walking controller for a \textsc{nao} robot.
Our empirical results show the benefit of modelling stochasticity when grounding a simulator for transfer to a stochastic real world environment.

\subsection{Cliff Walking}
\label{cliffwalkingdomainsection}

\begin{figure}[!tb]
    \centering
    \includegraphics[width=\columnwidth]{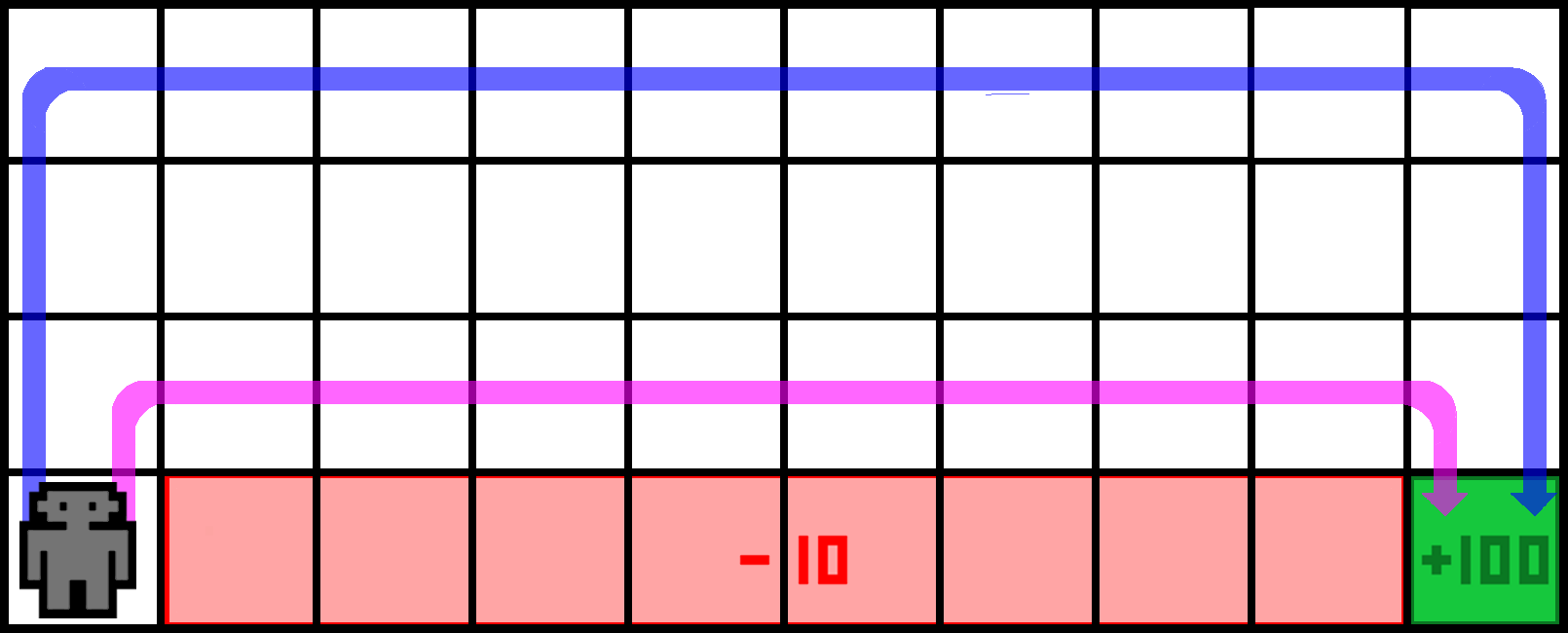}
    \caption{The agent starts in the bottom left and must reach the goal in the bottom right. Stepping into the red region penalizes the robot and ends the episode. The purple path is the most direct, but the blue path is safer when the transitions are stochastic.}
    \label{fig:gridworld_diagram}
\end{figure}

We verify the benefit of \textsc{sgat} using a classical reinforcement learning domain, the Cliff Walking grid world shown in Fig. \ref{fig:gridworld_diagram}. In this domain, an agent must navigate around a cliff to reach a goal. The episode terminates when it either reaches the goal (reward of $+100$) or falls into the cliff (reward of $-10$). There is also a small time penalty ($-0.1$ per time step), so the agent is incentivized to find the shortest path. There is no discounting, so the agent's objective is to maximize the sum of rewards over an episode. The agent can move up, down, left, or right. If it tries to move into a wall, the action has no effect. In our version of the problem, we assume we have a deterministic simulator, but in the ``real" environment, there is a small chance at every time step that the agent moves in a random direction instead of the direction it chose.

\begin{figure}[tb]
    \centering
    \includegraphics[width=\columnwidth]{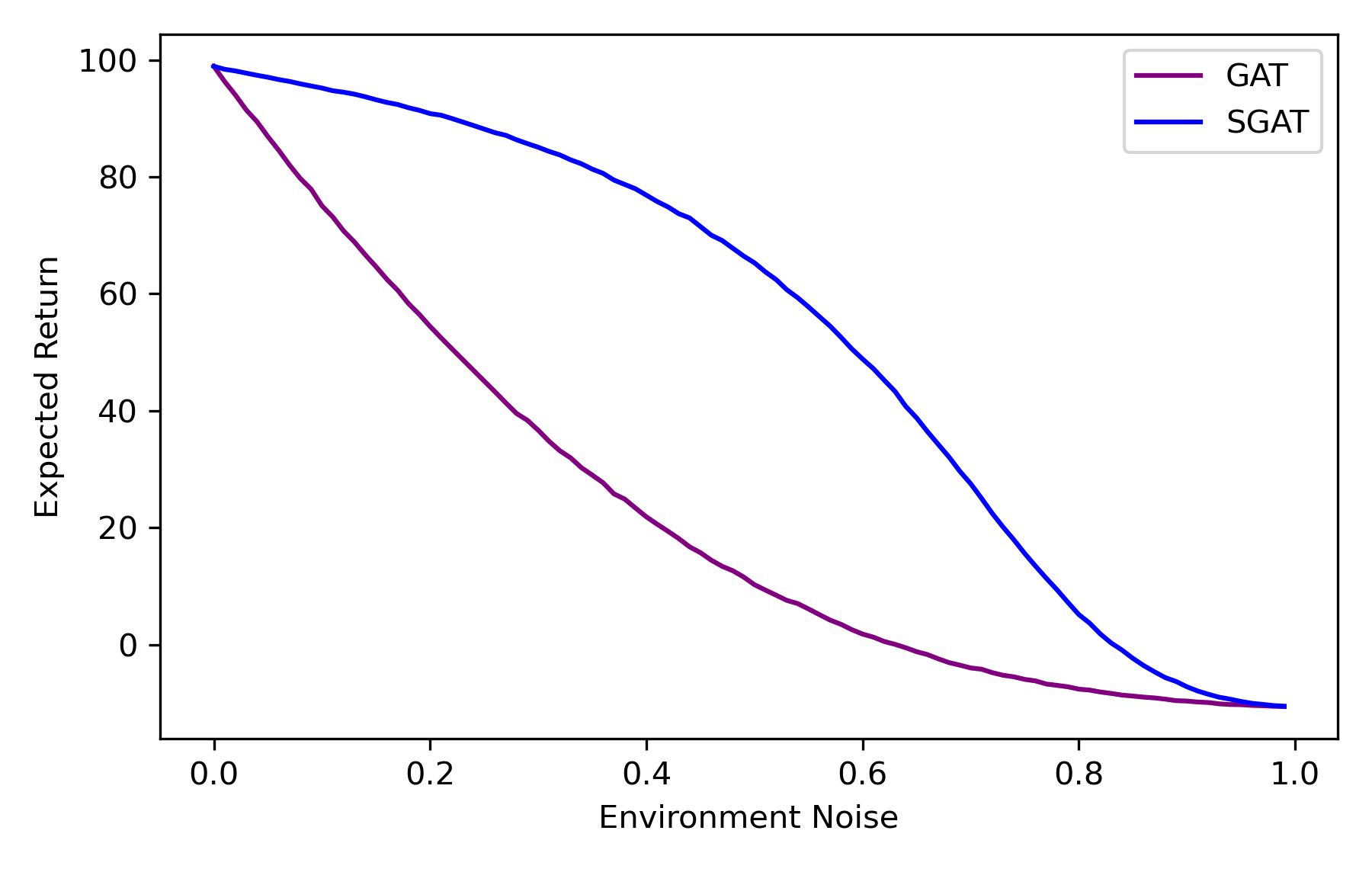}
    \caption{The y-axis is the average performance of a policy evaluated on the ``real" domain. The x-axis is the chance at each time step for the transition to be random. \textsc{sgat} outperforms \textsc{gat} for any noise value. Error bars not shown since standard error is smaller than 1 pixel.}
    \label{fig:grid_plot}
\end{figure}

Fig. \ref{fig:grid_plot} shows \textsc{gat} and \textsc{sgat} evaluated for different values of the environment noise parameter. Both the grounding steps and policy improvement steps (using policy iteration \cite{Suttonbook}) are repeated until convergence for both algorithms. To evaluate the resulting policy, we estimate the expected return from averaging 10,000 episodes. At a value of zero, the ``real" environment is completely deterministic. At a value of one, every transition is random. Thus, at both of these endpoints, there is no distinction between the expected return gained by the two algorithms.

For every intermediate value, \textsc{sgat} outperforms \textsc{gat}.
The policy trained using \textsc{gat} is unaware of the stochastic transitions, so it always takes the shortest and most dangerous path. Meanwhile the \textsc{sgat} agent learns as if it were training directly on the real environment in the presence of stochasticity.

\begin{figure}[tb]
    \centering
    \includegraphics[width=\columnwidth]{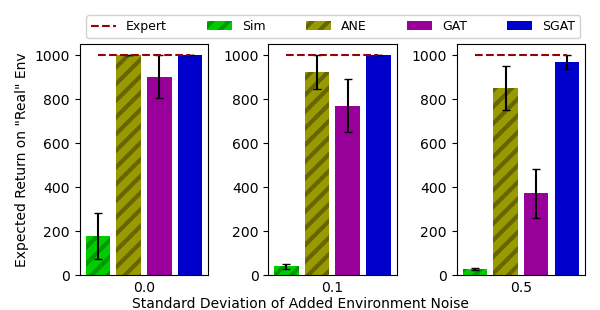}
    \caption{Sim-to-NoisyReal experiment on \textbf{InvertedPendulum}. The ``real" pendulum is 10 times heavier than the sim pendulum and has added Gaussian noise of different values. Error bars show standard error over ten independent training runs. Algorithms with striped bars used no real world data during training. \textsc{sgat} performs comparatively better in noisier target environments.}
    \label{fig:invp_s2r}
\end{figure}

\subsection {MuJoCo Domains}

\begin{figure*}[!tb]
\begin{subfigure}{\columnwidth}
    \includegraphics[width=\columnwidth]{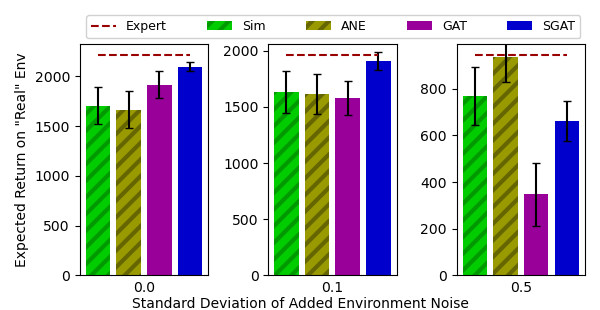}
    \caption{Sim-to-NoisySim}
    \label{fig:hc_s2s}
\end{subfigure}
\begin{subfigure}{\columnwidth}
\includegraphics[width=\columnwidth]{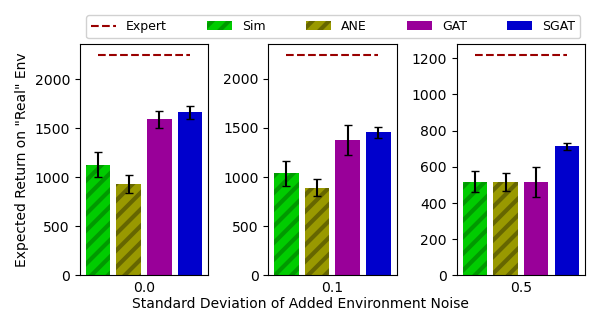}
    \caption{Sim-to-NoisyReal}
    \label{fig:hc_s2r}
\end{subfigure}
\caption{Sim-to-NoisySim and Sim-to-NoisyReal experiments on \textbf{HalfCheetah}. In the NoisyReal environment, the ``real" HalfCheetah's mass is 43\% greater than the sim HalfCheetah. Error bars show show standard error over ten independent training runs. Algorithms with striped bars used no real world data during training. When the ``real" environment is highly stochastic, \textsc{sgat} performs better than \textsc{gat}. Meanwhile, \textsc{ane} does poorly on less noisy scenarios.
}
\label{hc_exps}
\end{figure*}

Having shown the efficacy of \textsc{sgat} in a tabular domain, we now evaluate its performance in continuous control domains that are closer to real world robotics settings. 
We perform experiments on the OpenAI Gym MuJoCo environments to compare the effectiveness of \textsc{sgat} and \textsc{gat} when there is added noise in the target domain. We consider the case with just added noise and the case with both noise and domain mismatch between the source and target environments. We call the former Sim-to-NoisySim and the latter Sim-to-NoisyReal.
We chose the \textit{InvertedPendulum} and \textit{HalfCheetah} domains to test \textsc{sgat} in environments with both low and high dimensional state and action spaces. For policy improvement, we use an implementation of Trust Region Policy Optimization (\textsc{trpo}) \cite{TRPO}, from the stable-baselines repository \cite{stable-baselines}, using their hyperparameters for the respective domains.

We also compare against the action-noise-envelope (\textsc{ane}) approach \cite{Jakobi1995}; however, it is not a perfectly fair comparison in the sense that robustness approaches are sensitive to user-defined hyperparameters that predict the variation in the environment---in this case, the magnitude of the added noise. In grounding approaches, these parameters can be automatically learned from real world data. In these experiments, we chose the ANE hyperparameters (e.g., noise value) with a coarse grid search.

We simulate stochasticity in the target domains by adding Gaussian noise with different standard deviation values to the actions input into the environment.
We omit the results of Sim-to-NoisySim experiments for \textit{InvertedPendulum} because both algorithms performed well on the transfer task. Fig. \ref{fig:invp_s2r} shows the performance on the ``real" environment of policies trained four ways---naively on the ungrounded simulator, with \textsc{sgat}, with \textsc{gat}, and with \textsc{ane}.
In this Sim-to-NoisyReal experiment, \textsc{sgat} performs much better than \textsc{gat} when the stochasticity in the target domain increases.
Fig. \ref{hc_exps} shows the same experiment on \textit{HalfCheetah}, both with and without domain mismatch. Both these environments have an action space of $[-1, 1]$.

The red dashed lines show the performance of a policy trained directly on the ``real" environment until convergence, approximately the best possible performance.
The axes are scaled respective to this line.
The error bars show the standard error across 10 trials with different initialization weights. As the stochasticity increases, \textsc{sgat} policies perform better than those learned using \textsc{gat}. Meanwhile, \textsc{ane} does well only for particular noise values, depending on its training hyperparameters.

\subsection{\textsc{Nao} Robot Experiments}

Until this point in our analysis, we have used a modified version of the simulator in place of the “real” world so as to isolate the effect of stochasticity (as opposed to domain mismatch).  However, the true objective of this research is to enable transfer to real robots, which may exhibit very different noise profiles than the simulated environments. Thus, in this section, we validate \textsc{sgat} on a real humanoid robot learning to walk on uneven terrain.

Our robot experiments were conducted on the SoftBank \textsc{nao} bipedal robot, used in the RoboCup robot soccer competitions. For learning in simulation, we use the SimSpark physics simulator, used in the RoboCup 3D Simulation league. We compared \textsc{gat} and \textsc{sgat} by independently learning control policies using these algorithms to walk on uneven terrain, as shown in Fig. \ref{fig:lumpy_field_setup}. To create uneven ground in the lab, we placed foam packing material under the turf of the robot soccer field. On this uneven ground, the walking dynamics become more random, since the forces acting on the foot are slightly different every time the robot takes a step. For the walk policy, we use the rUNSWift walk engine and the initial published parameter set $\theta_0$ \cite{walkengine2014}. This initial unoptimized policy achieves a speed of $14.66 \pm 1.65$ cm/s on the uneven terrain. The walk engine contains 16 parameters that we optimize on the grounded simulator using Covariance Matrix Adaptation - Evolutionary Strategy (\textsc{cma-es}), with a population size of 150. Each trajectory lasts for 7.5 seconds on the simulator or terminates when the robot falls down. The reward function is a sum of the forward velocity of the robot (in cm/s) and an early termination penalty of -15 if the robot falls down. Thus, \textsc{cma-es} optimizes the policy for walks that are faster and more stable in the grounded simulator. 

On flat ground, both methods produced very similar policies, but on the uneven ground, the policy learned using \textsc{sgat} was more successful than a policy learned using \textsc{gat}.
We performed ten trial runs of the best policy learned using \textsc{sgat} and \textsc{gat} after each grounding step, and the average speed of the robot on the uneven terrain is shown in Table \ref{table:nao}.
The policy learned using \textsc{sgat} takes shorter steps and stays upright, thereby maintaining its balance on the uneven terrain, whereas the policy produced using \textsc{gat} learned to lean forward and walk faster, but fell down more often due to the uneven terrain. This result is best visualized in the supplementary video attached with this paper. Both algorithms produce policies that improve the walking speed across grounding steps. The \textsc{gat} policy after the second grounding step always falls over, whereas the \textsc{sgat} policy was more stable and finished the course 9 out of 10 times.

\begin{table}[!tb]
\caption{Speed and stability of \textsc{nao} robot walking on uneven ground. The initial policy $\theta_0$ walks at $14.66 \pm 1.65$ cm/s and always falls down. Both \textsc{sgat} and \textsc{gat} find policies that are faster, but \textsc{sgat} policies are more stable than policies learned using \textsc{gat}.}
\centering
\label{table:nao}
\begin{tabular}{| c | c | c | c | c |}
    \hline
    & \multicolumn{2}{|c|}{Grounding Step 1} & \multicolumn{2}{|c|}{Grounding Step 2} \\
    \hline
    & Speed (cm/s) & Falls & Speed (cm/s) & Falls \\
    \hline
    \textsc{gat} & $15.7 \pm 2.98$ & 6/10 & $18.5 \pm 3.63$ & 10/10 \\
    \hline
    \textsc{sgat} & $16.9 \pm 0.678$ & \textbf{0/10} & $18.0 \pm 2.15$ & \textbf{1/10} \\
    \hline
\end{tabular}
\end{table}

\section{DISCUSSION AND LIMITATIONS}

In our experiments, we observe that stochastic transitions can have a detrimental effect on how \textsc{gat} grounds the simulator; however both algorithms perform similarly on deterministic environments. In real world scenarios, we cannot know how stochastic an environment is before testing. This fact suggests that we should default to \textsc{sgat}.

In this work, we have only considered deterministic simulators, but simulators may have stochastic transitions as well, especially if the simulator was designed to anticipate process noise. However, when using an action transformer grounding approach, stochastic simulators make the learning problem more difficult. We can no longer sample from the distribution provided by the forward model. Instead, the inverse model must take in a distribution over states and output a distribution over actions.


\section{CONCLUSION AND FUTURE WORK}

In this work, we introduced  \textit{Stochastic Grounded Action Transformation} (\textsc{sgat}), a sim-to-real algorithm, which gracefully handles policy learning in simulation when the target domain is stochastic.

First, in \textit{Cliff Walking}, we empirically showed that by accounting for target domain stochasticity, \textsc{sgat} policies can produce much better average returns on the real environment than \textsc{gat} policies. Then, we evaluated performance on two MuJoCo environments and showed that \textsc{sgat} outperforms other algorithms when the target environment is highly stochastic. To verify the algorithm in a real world transfer setting, we trained a \textsc{nao} humanoid robot to walk over uneven terrain. The empirical results show that \textsc{sgat} policies are more robust to process noise and walk more robustly over the uneven terrain compared to \textsc{GAT} and hand-coded polices.

Though we have focused on GAT, similar distributional approaches are directly applicable to other black-box sim-to-real techniques \cite{rgat, higuera17}, and possibly to some white-box techniques as well. Many sim-to-real algorithms are designed to find correct environment parameters, but the most accurate model might be one in which these parameters are sampled from a distribution. Investigating this scenario is a promising direction for future work.

\section{ACKNOWLEDGMENT}
\small{This work has taken place in the Learning Agents Research Group (LARG) at UT Austin.  LARG research is supported in part by NSF (CPS-1739964, IIS-1724157, NRI-1925082), ONR (N00014-18-2243), FLI (RFP2-000), ARL, DARPA, Lockheed Martin, GM, and Bosch.  Peter Stone serves as the Executive Director of Sony AI America and receives financial compensation for this work.  The terms of this arrangement have been reviewed and approved by the University of Texas at Austin in accordance with its policy on objectivity in research.}

\bibliographystyle{IEEEtran}
\bibliography{mybib}

\end{document}